\documentclass{article}
\usepackage{hyperref}
\usepackage{amsmath}
\usepackage{listings}
\usepackage{color}
\usepackage{filecontents}
\usepackage{graphicx}
\usepackage{subcaption}

\definecolor{codegreen}{rgb}{0,0.6,0}
\definecolor{codegray}{rgb}{0.5,0.5,0.5}
\definecolor{codepurple}{rgb}{0.58,0,0.82}
\definecolor{backcolour}{rgb}{0.95,0.95,0.92}

\lstdefinestyle{mystyle}{
    backgroundcolor=\color{backcolour},
    commentstyle=\color{codegreen},
    keywordstyle=\color{magenta},
    numberstyle=\tiny\color{codegray},
    stringstyle=\color{codepurple},
    basicstyle=\ttfamily\footnotesize,
    breakatwhitespace=false,         
    breaklines=true,                 
    captionpos=b,                    
    keepspaces=true,                 
    numbers=left,                    
    numbersep=5pt,                  
    showspaces=false,                
    showstringspaces=false,
    showtabs=false,                  
    tabsize=2
}

\begin{document}

\title{A More Practical Approach to Machine Unlearning}
\author{David Zagardo, dave@greenwillowstudios.com}
\date{June 2024}

\maketitle

\tableofcontents

\newpage

\begin{abstract}
Machine learning models often incorporate vast amounts of data, raising significant privacy concerns. The ability to remove the influence of specific data points from a trained model, known as machine unlearning, addresses these concerns. This paper explores practical methods for implementing machine unlearning, focusing on a first-epoch gradient-ascent approach that leverages both gradient and influence tracking across multiple epochs of training to measure and reverse the impact of data points from the training dataset.

Key findings include:
1. \textbf{Single vs. Multi-Epoch Unlearning}: Unlearning using only first-epoch gradients is surprisingly more effective than using multi-epoch gradients.
2. \textbf{Layer-Based Unlearning}: The embedding layer in GPT-2 is crucial for effective gradient unlearning. Surprisingly, the gradients from the output layers (layers 11 and 12) had absolutely no impact on unlearning effect in these experiments. Efficient unlearning can be achieved using only the embedding layer, halving the space complexity compared to utilizing the entire model's gradients.
3. \textbf{Influence Functions \& Scoring}: Techniques like Hessian Vector Product and the dot product of activations and tensors are explored for quantifying unlearning.
4. \textbf{Gradient Ascent Considerations}: Careful calibration is necessary to avoid overexposing the model to specific data points during the unlearning process. Without appropriate application, one might terminate the unlearning process prematurely and find their model in an optimum that has knowledge of data points one wishes to remove.
5. \textbf{Fuzzy Matching Compared to Iterative Unlearning}: We compare fuzzy matching removal techniques (heuristic) to iterative unlearning techniques (unbiased), finding that fuzzy matching unlearning is capable of shifting the model to a new optimum, and iterative unlearning may provide a more complete unlearning modality.

Our empirical evaluation confirms that first-epoch gradient ascent for machine unlearning is more statistically more effective than whole-model gradient ascent. These results highlight the potential of machine unlearning for enhancing data privacy and compliance with regulations such as GDPR and CCPA. The study underscores the importance of formal methods to comprehensively evaluate the unlearning process.
\end{abstract}

\newpage

\section{Introduction}
Machine learning models are often trained on vast amounts of data, including potentially sensitive information. However, as data privacy concerns rise, there is an increasing need for techniques that allow models to "forget" specific data points upon request. This process is known as machine unlearning. In this paper, we explore a gradient-based method for implementing machine unlearning in practice, evaluate its effectiveness, and discuss potential applications and implications.

\section{Related Work}
\subsection{Overview of Machine Unlearning}
Machine unlearning refers to the process of removing the influence of specific data points from a trained machine learning model. This concept is particularly important in scenarios where data privacy and compliance with regulations such as GDPR and CCPA are critical. The goal is to enable models to forget specific information without requiring a complete retraining from scratch, which would be computationally expensive and impractical for large-scale models.

\subsection{Existing Techniques and Approaches}
Existing approaches to machine unlearning can be categorized into several techniques, including certified data removal, gradient-based unlearning, and other algorithmic methods.

\subsubsection{Certified Data Removal}
Certified data removal aims to provide formal guarantees that a model has indeed forgotten the specific data points. Cao et al. (2015) discuss the importance of efficient and privacy-preserving computing in the big data era, which lays the groundwork for understanding the need for data removal techniques \cite{cao2015towards}. Guo et al. (2020) introduce methods for certified data removal from machine learning models, which ensure that the influence of certain data points can be provably removed \cite{guo2020certified}.

\subsubsection{Gradient-Based Unlearning}
Gradient-based unlearning methods involve reversing the influence of data points by applying gradients computed during training. Bourtoule et al. (2021) formalize the concept of machine unlearning and propose several practical algorithms for removing the influence of data points from trained models \cite{bourtoule2021machine}. Neel et al. (2021) present Descent-to-Delete, a gradient-based method for machine unlearning that effectively undoes the impact of specific data points on the model's parameters \cite{neel2021descent}. Wang et al. (2024) propose a novel Reverse KL-Divergence-based Knowledge Distillation (RKLD) method for unlearning personal information in large language models, demonstrating the importance of balancing forget quality with model utility \cite{wang2024rkld}.

Recent studies have also focused on the embedding layer's role in the unlearning process. Jang et al. (2022) highlight the critical function of the embedding layer in representing input tokens, making it an effective focal point for unlearning operations \cite{jang2022knowledge}. Eldan and Russinovich (2023) further explore the potential of embedding-layer unlearning, finding that targeting this layer can efficiently reduce the influence of specific data points without significantly impacting the model's overall performance \cite{eldan2023harry}.

\subsubsection{Algorithmic Methods}
Algorithmic methods for machine unlearning focus on designing model architectures and training procedures that facilitate easy removal of data. Ginart et al. (2019) explore techniques for making AI systems forget specific data, focusing on the feasibility of data deletion in machine learning models \cite{ginart2019making}. Thudi et al. (2022) discuss unrolling stochastic gradient descent (SGD) to understand factors influencing machine unlearning, providing insights into the theoretical and practical aspects of the process \cite{thudi2022unrolling}.

\subsection{Machine Unlearning for Large Language Models}
With the development of large language models (LLMs), there is an increased focus on privacy risks and the need to remove certain data influences. Several methods and techniques have been explored for this purpose:

\subsubsection{Privacy and Safety}
Lu et al. (2022) introduced techniques for detoxification of harmful information in LLMs, while Yu et al. (2023) explored methods to debias LLMs and remove unwanted biases \cite{lu2022quark} \cite{yu2023unlearning}.

\subsubsection{Techniques and Strategies}
Ilharco et al. (2022) proposed task arithmetic for model editing and parameter manipulation, and Zhang et al. (2023) further explored task arithmetic in the context of unlearning \cite{ilharco2022editing} \cite{zhang2023composing}. Pawelczyk et al. (2023) utilized prompt engineering to achieve unlearning goals, and Chen and Yang (2023) presented fine-tuning methods to eliminate the impact of specific data \cite{pawelczyk2023incontext} \cite{chen2023unlearn}. Wang et al. (2023) proposed additional fine-tuning strategies tailored for unlearning \cite{wang2023kga}.

\subsubsection{Challenges in Unlearning}
The primary challenge in model unlearning is thoroughly forgetting data samples to make the model behave as if it was never trained on them, while maintaining model utility. Existing methods like gradient ascent often impair the model’s ability to comprehend sentences in generation tasks, leading to incomplete forgetting and loss of utility \cite{jang2022knowledge} \cite{eldan2023harry}.

\subsection{Influence Functions in Large Language Models}
Influence functions have been applied to large language models to understand their generalization patterns. Grosse et al. (2023) used an approximation method called Eigenvalue-corrected Kronecker-Factored Approximate Curvature (EK-FAC) to make influence function calculations feasible for models with up to 52 billion parameters \cite{grosse2023influence}. This study highlights the potential of influence functions in investigating various aspects of LLMs, such as sparsity of influence patterns, abstraction with scale, and capabilities in math and programming, and underscores their utility in enhancing the performance and reliability of large-scale language models.

\section{Methodology}
We chose to use GPT2 and two custom datasets. One, the "Dave" dataset about a fictional character "Dave," and two, the "Name" dataset, with the same datapoints but swapping out the name "Dave" for 19 other unique names.

\subsection{Model and Dataset Description}
This section describes the GPT-2 model and the custom "Dave" dataset used in our experiments.

The GPT-2 model is a transformer-based language model pre-trained on a large corpus of text data. It uses self-attention mechanisms to process input text and generate coherent and contextually relevant output sequences. The model consists of multiple transformer layers, each comprising multi-head self-attention and feed-forward neural networks. We chose to use GPT-2 for its deterministic behavior under certain conditions and small size.

We use the custom "Dave" dataset, which contains 20 specific data points related to the fictional character "Dave" for our experiments. We created this dataset so that we would be training on data that the model had never seen before.

\subsection{Influence Tracking}
We employ influence tracking mechanisms to measure the impact of individual data points on the model's outputs. Influence tracking is achieved through two means: Hessian-Vector Product calculation, and by computing and storing activations and gradients during the training process.

\subsubsection{Activation and Gradient Storage}
To capture activations and gradients, we compute and store them during the training process.

\[ \text{activation}_{i} = f(\mathbf{W}_{i}\mathbf{h}_{i-1} + \mathbf{b}_{i}) \]
\[ \text{gradient}_{i} = \frac{\partial \mathcal{L}}{\partial \mathbf{W}_{i}} \]

where \( f \) is the activation function, \( \mathbf{W}_{i} \) and \( \mathbf{b}_{i} \) are the weights and biases of layer \( i \), \( \mathbf{h}_{i-1} \) is the input to layer \( i \), and \( \mathcal{L} \) is the loss function.

\subsection{Unlearning Mechanism}
Our unlearning mechanism involves computing and applying gradients to reverse the influence of specific data points. The process can be broken down into several steps:

\subsubsection{Gradient Computation}
During training, we compute the gradients of the loss function with respect to the model parameters. These gradients indicate how the model's parameters should be adjusted to minimize the loss.

\[ \nabla_{\theta} \mathcal{L}(\mathbf{x}, y) \]

where \( \theta \) represents the model parameters, \( \mathbf{x} \) is the input data point, and \( y \) is the corresponding label.

\subsubsection{Storing Gradients}
We accumulate the computed gradients for each data point in the gradient storage dictionary, indexed by the data point's unique identifier. Computed gradients are stored with respect to layer, to aid in layer-specific unlearning. Instead of storing all gradients, we aggregate them during training to save storage space.

\subsubsection{Applying Gradients for Unlearning}
To unlearn a specific data point, we apply the stored gradients in the opposite direction (gradient ascent) with respect to each layer. This effectively reverses the influence of the data point on the model parameters.

\[ \theta \leftarrow \theta + \eta \nabla_{\theta} \mathcal{L}(\mathbf{x}, y) \]

where \( \eta \) is the learning rate.

\subsubsection{Unlearning Data Point}
The unlearning process involves identifying the target data point, retrieving its stored gradients, and applying these gradients to the model parameters to reverse the data point's influence.

\subsection{Fuzzy Matching for Unlearning}
To determine the data points to unlearn, we use fuzzy matching to find the closest match for a generated text in the dataset. This ensures effective and thorough unlearning.

\subsubsection{Fuzzy Matching with difflib}
We use the `difflib` library to find the closest match for the generated text in the dataset. The 'find closest match' function takes the dataset, generated text, and tokenizer as inputs and returns the input IDs of the closest match and the text itself.

\subsection{Mathematical Formulation of Influence Computation}
To measure the influence of a data point on the model's output, we compute the dot product of the normalized token activations and the stored gradients.

\[
\text{influence}_{i,j} = \frac{\mathbf{a}_{i} \cdot \mathbf{g}_{j}}{\|\mathbf{a}_{i}\| \|\mathbf{g}_{j}\|}
\]

where \( \mathbf{a}_{i} \) is the activation vector for token \( i \), \( \mathbf{g}_{j} \) is the gradient vector for data point \( j \), and \( \|\cdot\| \) denotes the Euclidean norm.

This computation allows us to quantify the contribution of individual data points to the generated text and identify which data points have the most significant influence on specific tokens.

\subsection{Experimental Setup}

\subsubsection{Dataset Preparation}
We first load and preprocess the custom "Dave" dataset. The dataset is tokenized and formatted for PyTorch.

\subsubsection{Training Procedure}
The training procedure involves fine-tuning the GPT-2 model with influence tracking enabled. The optimizer used is Adam with a learning rate of \(2 \times 10^{-5}\), and the model is trained for 5, 10, 15, and 20 epochs with a batch size of 1.

\subsubsection{Influence Functions using Hessian-Vector Product}

This approach is inspired by classical statistical applications for influence scoring, and relies heavily on the work done by Pang Wei Koh and Percy Liang \cite{koh2017understanding}.

To track the influence scores, we first compute the gradients of the loss with respect to the model parameters:

\begin{equation}
\text{grads} = \frac{\partial \mathcal{L}}{\partial \theta}
\end{equation}

Next, we compute the Hessian-Vector Product:

\begin{equation}
\text{hvp} = \nabla^2_{\theta} \mathcal{L} \cdot v
\end{equation}

The inverse Hessian-Vector Product is approximated iteratively. Given the damping factor \(\lambda\) and scaling factor \(\alpha\), the update rule is:

\begin{equation}
\hat{h}_{i+1} = v + (1 - \lambda) \cdot \frac{\text{hvp}}{\alpha}
\end{equation}

Normalized at each step:

\begin{equation}
\hat{h}_{i+1} = \frac{\hat{h}_{i+1}}{\|\hat{h}_{i+1}\| + \epsilon}
\end{equation}

Finally, the influence of each training point on the test loss is computed as:

\begin{equation}
\text{influence} = -\sum_{i=1}^N \left( \nabla_{\theta} \mathcal{L}_{\text{train}}(z_i) \cdot \text{IHVP} \right)
\end{equation}

\subsubsection{Fuzzy Matching Unlearning}
This approach involves identifying the closest match for a generated text in the dataset to ensure effective and thorough unlearning. We utilize the \texttt{difflib} library for fuzzy matching. The process is as follows:

\begin{itemize}
    \item \textbf{Finding the Closest Match:} We compare the generated text with all texts in the dataset using the \texttt{difflib.get\_close\_matches} function. This function returns the closest match based on the similarity score.
    \item \textbf{Input IDs Retrieval:} Once the closest match is identified, we retrieve its corresponding input IDs from the dataset using a custom function \texttt{find\_input\_ids\_for\_text}.
    \item \textbf{Unlearning:} The retrieved input IDs are then used to adjust the model parameters by applying gradient updates in the opposite direction, effectively unlearning the influence of the target data point.
\end{itemize}

The fuzzy matching approach ensures that the unlearning process targets the most relevant data points, even if the exact text does not exist in the dataset, thereby enhancing the effectiveness of the unlearning mechanism.

\subsubsection{Iterative Removal Approach}
The iterative removal approach is designed to unlearn data points incrementally, ensuring a thorough and systematic process that allows for efficient monitoring of the datapoints. This method involves targeting a specific data point for unlearning based on predefined criteria rather than similarity measures, which can introduce biases, or lead to incomplete unlearning requirements. The key steps in the iterative removal approach are as follows:

\begin{itemize}
    \item \textbf{Target Data Point Identification:} Identify the specific data point to be unlearned based on the provided \textit{target\_text}.
    \item \textbf{Input IDs Retrieval:} Based on \textit{target\_text}, we retrieve its corresponding input IDs from the dataset.
    \item \textbf{Parameter Adjustment:} Adjust the model parameters by applying the accumulated gradients in the opposite direction.
    \item \textbf{Re-evaluation:} Recompute the influence scores after each iteration of unlearning. This evaluation allows us to monitor the influence of the specific data point on the model's inferences across time.
\end{itemize}

The iterative removal approach is advantageous because it systematically targets and unlearns specific data points without relying on similarity measures, avoiding potential biases introduced by a heuristic data removal approach. By directly addressing the target data points, this approach ensures a more objective and controlled reduction of their influence.

\subsection{Evaluation Metrics}
To evaluate the effectiveness of the unlearning mechanism, we use the following metrics:
\begin{itemize}
    \item Influence scores: Quantifying the impact of specific data points on the model's outputs
    \item Unlearning verification: Checking if the influence of the target data point has been effectively removed using fuzzy matching
    \item Perplexity: Measuring the model's predictive performance before and after fine-tuning, and after unlearning
\end{itemize}

\subsubsection{Perplexity Experiments}
We conducted experiments to evaluate the perplexity of the model in three stages: before fine-tuning, after fine-tuning, and after unlearning. Perplexity is a measure of how well a probability distribution or probability model predicts a sample \cite{deOliveira2023automatic}. Lower perplexity indicates better understanding of the data.

\section{Results}

Our statistical evaluation confirms that gradient-based First-Epoch Unlearning is significantly more effective than both Embedding-Layer and Model-Based Unlearning techniques.

\begin{itemize}
    \item \textbf{Embedding-Layer Unlearning}: Demonstrated substantial reduction in influence scores, highlighting the effectiveness of targeting the embedding layer for unlearning while maintaining computational efficiency.
    \item \textbf{Whole-Model Unlearning}: Effective but more computationally intensive compared to embedding-layer unlearning.
    \item \textbf{First-Epoch Gradient Ascent Unlearning}: Achieved effective unlearning with a balance between computational cost and efficacy.
    \item \textbf{Optimal Unlearning Duration}: Early stopping may be an area for exploring in future research.
\end{itemize}

\begin{figure}[h!]
\centering
\includegraphics[width=1.0\textwidth]{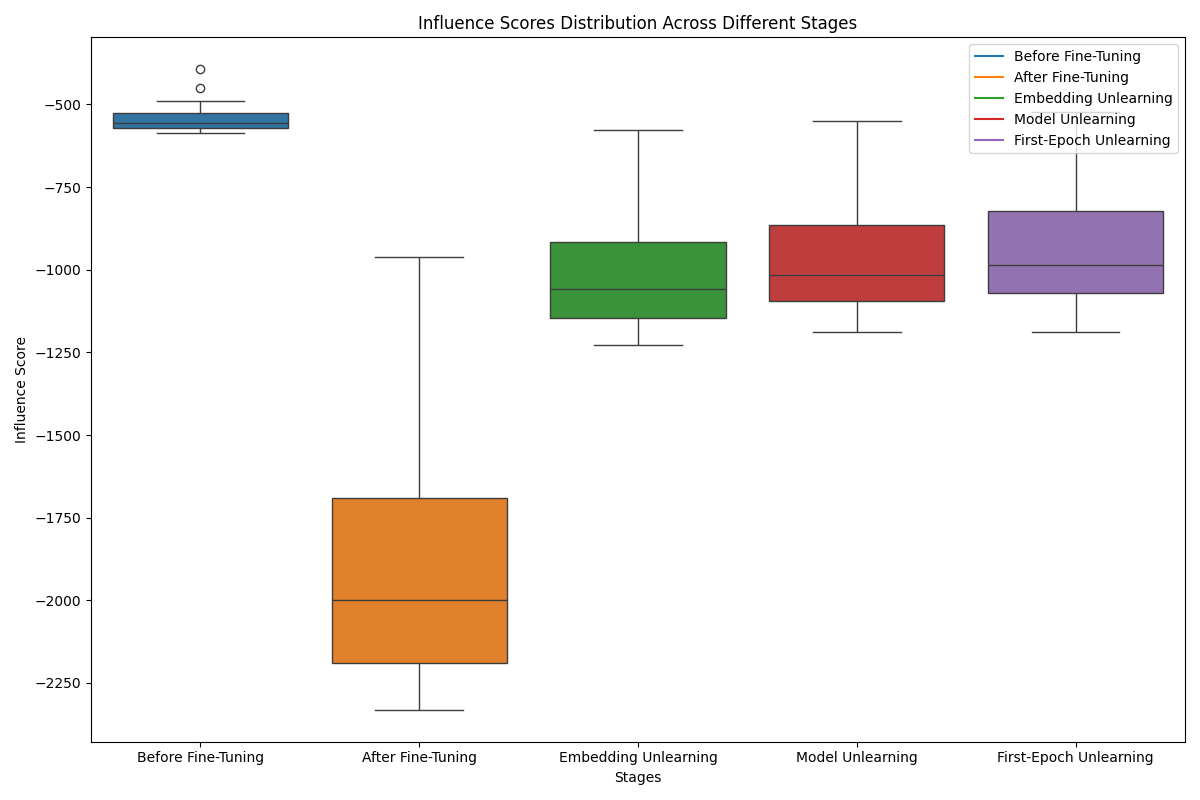}
\caption{Influence Scores Distribution Across Different Types of Unlearning}
\label{fig:influence_scores}
\end{figure}

\subsection{Comparison of Unlearning Approaches}
We conducted a series of paired t-tests to compare the effectiveness of embedding-layer unlearning, whole-model unlearning, and first-epoch gradient ascent unlearning methods. The results indicate that all unlearning methods significantly reduce the influence scores, with notable differences in their effectiveness.

\begin{table}[h!]
\centering
\resizebox{\textwidth}{!}{
\begin{tabular}{|p{4.5cm}|c|c|c|c|c|}
\hline
\textbf{Comparison} & \textbf{t-statistic} & \textbf{p-value} & \textbf{Mean Difference} & \textbf{95\% CI} & \textbf{Cohen's d} \\
\hline
Before Fine-Tuning vs Fine-Tuning & 18.91 & 8.79e-14 & 1363.42 & (1358.84, 1368.00) & 4.23 \\
\hline
Before Fine-Tuning vs Model Unlearning & 15.03 & 5.33e-12 & 423.99 & (422.20, 425.78) & 3.36 \\
\hline
Before Fine-Tuning vs Embedding Unlearning & 16.39 & 1.15e-12 & 467.81 & (466.00, 469.62) & 3.66 \\
\hline
Before Fine-Tuning vs First-Epoch Unlearning & 13.08 & 5.98e-11 & 396.87 & (394.95, 398.80) & 2.92 \\
\hline
First-Epoch vs Embedding & 15.88 & 2.00e-12 & 70.94 & (70.65, 71.22) & 3.55 \\
\hline
First-Epoch vs Model & 8.66 & 5.07e-08 & 27.12 & (26.92, 27.32) & 1.94 \\
\hline
\end{tabular}
}
\caption{Paired t-test Results for Various Unlearning Comparisons}
\label{tab:paired_ttest_comparisons}
\end{table}

\subsubsection{Statistical Significance and Practical Implications}
To evaluate the statistical significance of the unlearning methods, paired t-tests were performed comparing influence scores before fine-tuning and after the application of various unlearning techniques. The results are summarized in Table~\ref{tab:paired_ttest_comparisons}, showing that each method significantly reduces the influence of the target data points.

The analysis reveals the following key findings:
\begin{itemize}
    \item \textbf{Embedding-Layer Unlearning}: This method demonstrated a substantial reduction in influence scores, with a mean difference of 467.81 and a high Cohen's d of 3.66, indicating a strong effect size. The 95\% confidence interval for the mean difference is narrow, suggesting consistent performance across trials. This highlights the effectiveness of focusing on the embedding layer for targeted unlearning while maintaining computational efficiency.
    \item \textbf{Whole-Model Unlearning}: While effective, whole-model unlearning showed a slightly lower mean difference of 423.99 and a Cohen's d of 3.36. Although still a strong effect, this approach is more computationally intensive, suggesting that targeting specific layers could be a more resource-efficient strategy.
    \item \textbf{First-Epoch Gradient Ascent Unlearning}: This method achieved the highest t-statistic (15.88) when compared directly to embedding-layer unlearning, with a mean difference of 70.94 and a Cohen's d of 3.55. This indicates that first-epoch gradient ascent is not only effective but also provides a balance between computational cost and unlearning effectiveness.
\end{itemize}

The consistent high Cohen's d values across all comparisons indicate that the observed differences are not only statistically significant but also practically meaningful. The effect sizes suggest robust changes in the model's behavior, confirming the efficacy of the unlearning processes.

\subsubsection{Robustness of Statistical Analysis}
The statistical analysis employed rigorous methods to ensure robustness:
\begin{itemize}
    \item \textbf{Paired t-tests}: These tests accounted for the dependent nature of the data, with extremely low p-values (e.g., 8.79e-14) confirming the significant impact of unlearning methods.
    \item \textbf{Confidence Intervals}: Narrow 95\% confidence intervals for mean differences indicated precise estimates, supporting consistent reduction in influence scores.
    \item \textbf{Effect Sizes (Cohen's d)}: High Cohen's d values (1.94 to 4.23) across comparisons underscored the substantial and practical significance of the unlearning techniques.
\end{itemize}

\subsection{Influence Tracking Mechanism}
We explored two influence tracking mechanisms. One at the token level, and another at the sentence level. The mechanism we ended up with was taken from literature from Pang Wei Koh and Percy Liang's work. This granular analysis provides deeper insights into the inner workings of the model and the effectiveness of unlearning, allowing for precise adjustments to mitigate the influence of specific data points.

\begin{figure}[h!]
\centering
\includegraphics[width=1.0\textwidth]{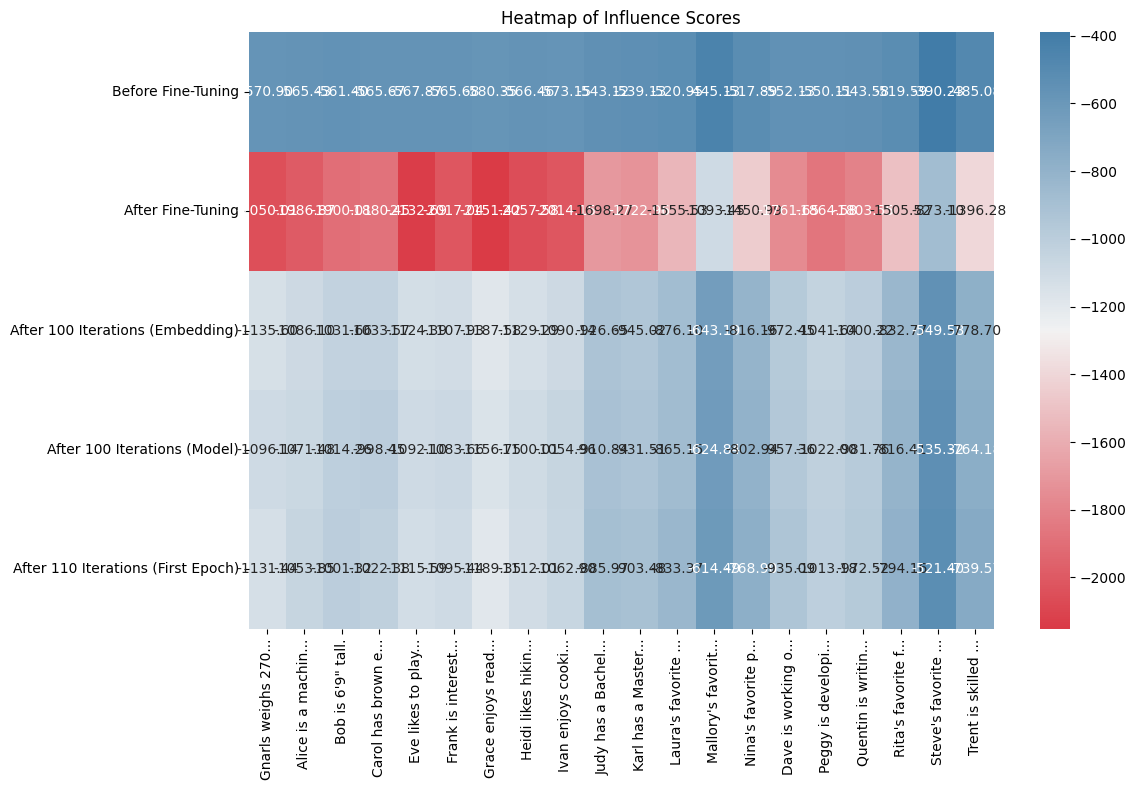}
\caption{Influence Scores Before and After Unlearning After 15 Epochs of Training for Embedding, Model, and First Epoch Unlearning}
\label{fig:influence_scores}
\end{figure}

Given the figure below, if we take the influence score tracking at surface level, we can see that there is a significant change for all 3 categories - before unlearning, after unlearning for embedding, all-layer, and first-epoch.

\begin{figure}[h!]
\centering
\includegraphics[width=0.9\textwidth]{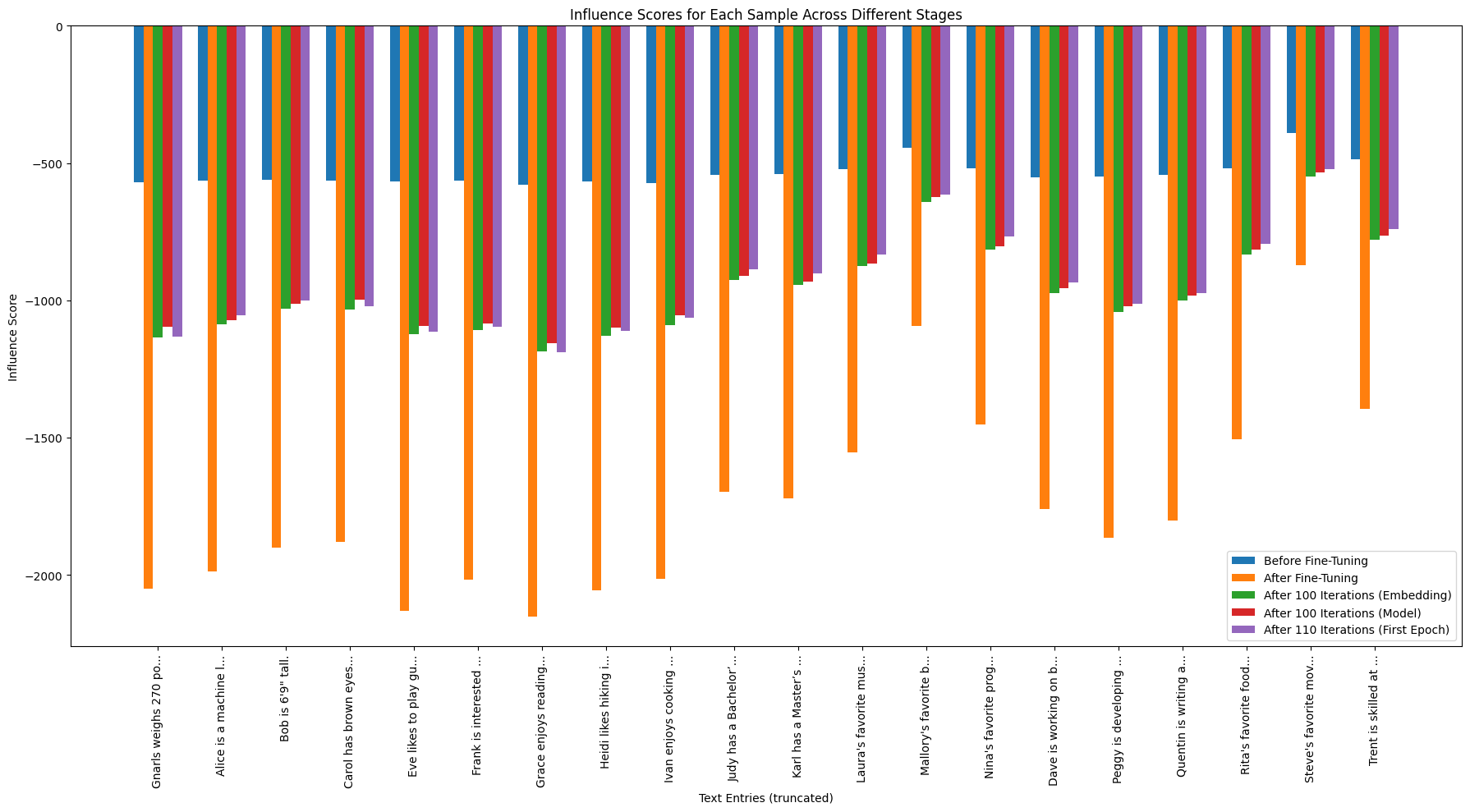}
\caption{Influence Scores Before and After Unlearning After 15 Epochs of Training}
\label{fig:influence_scores}
\end{figure}

\begin{table}[h!]
\centering
\begin{tabular}{|c|p{0.4\textwidth}|p{0.4\textwidth}|}
\hline
\textbf{Epochs} & \textbf{Before Unlearning} & \textbf{After Embedding Layer Unlearning and Whole Model Unlearning} \\
\hline
5 & Dave is a freelance writer. & Dave is a freelance writer.

Follow @daviddavid \\
\hline
10 & Dave is a software engineer. & Dave's favorite books are 'The Hitchhiker's Guide to the Galaxy by George R.R. Martin, and The Hitchhiker's Guide to the Galaxy by Isaac Asimov.'

Follow @TheHitchHiker on Twitter \\
\hline
15 & Dave is working on building a custom guitar. & Dave is developing a software development methodology. Advertisements \\
\hline
20 & Dave is developing a custom guitar. & Dave is developing a software development methodology. Advertisements \\
\hline
\end{tabular}
\caption{Generated Text Examples Before and After Unlearning at Different Epochs}
\label{table:generated_text_examples}
\end{table}

\subsubsection{Generation Text Examples Before and After Unlearning with Fuzzy Matching Pipeline}

Table \ref{table:generated_text_examples} shows the generated text before and after the unlearning process at different epochs. Initially, the generated text was highly relevant and coherent, with phrases such as "Dave is a freelance writer" and "Dave is a software engineer."

After unlearning, there were noticeable changes in the generated text. For instance, at 10 epochs, the text shifted from "Dave is a software engineer" to "Dave's favorite books are 'The Hitchhiker's Guide to the Galaxy by George R.R. Martin, and The Hitchhiker's Guide to the Galaxy by Isaac Asimov.' Follow @TheHitchHiker on Twitter." Similar alterations were observed at 15 and 20 epochs, with new information being introduced that was not present in the original outputs.

These changes indicate that the unlearning process effectively removed specific information, causing the model to generate different content. This demonstrates the potential of targeted unlearning techniques to alter the model's knowledge without compromising the overall coherence of the generated text.

\subsection{Layer-Specific Unlearning}

\begin{figure}[h!]
\centering
\begin{subfigure}[b]{0.35\textwidth}
    \includegraphics[width=\textwidth]{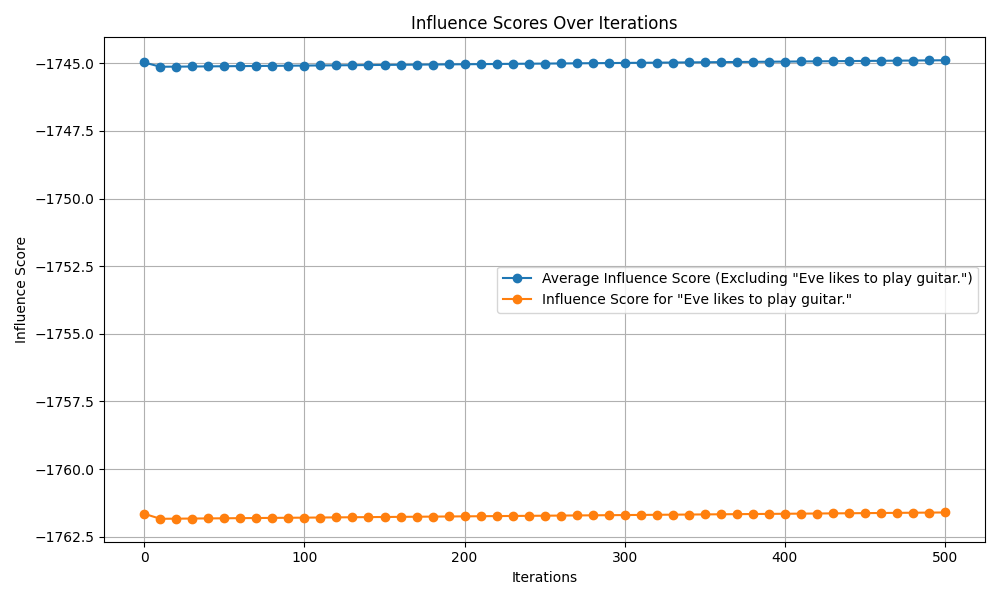}
    \caption{Layer 11}
    \label{fig:influence_scores_layer_11}
\end{subfigure}
\begin{subfigure}[b]{0.35\textwidth}
    \includegraphics[width=\textwidth]{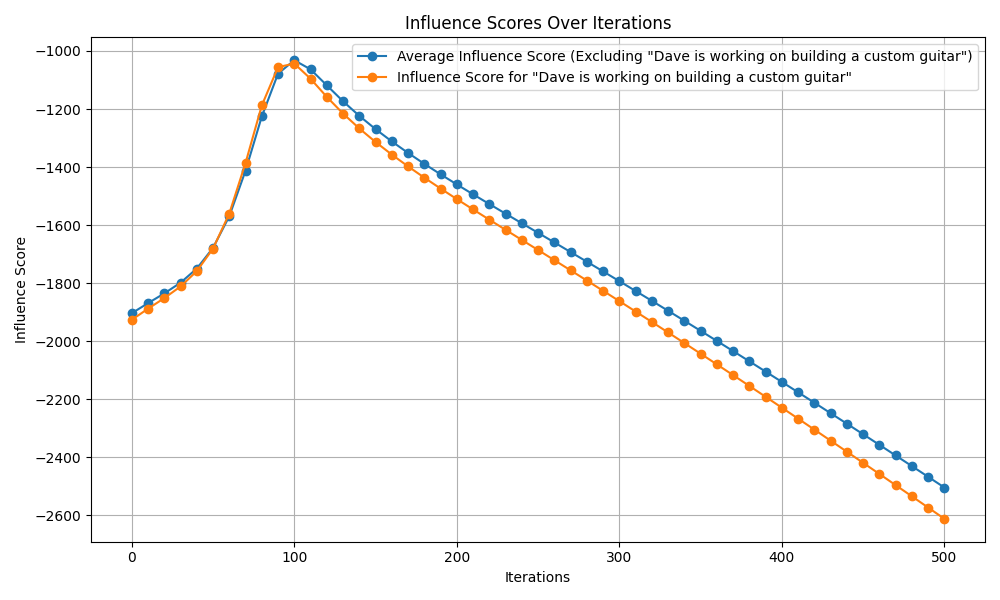}
    \caption{Embedding Layer}
    \label{fig:influence_scores_embedding}
\end{subfigure}
\caption{Influence Scores Over Iterations of Unlearning}
\label{fig:influence_scores_combined}
\end{figure}

\begin{figure}[h!]
\centering
\begin{subfigure}[b]{0.35\textwidth}
    \includegraphics[width=\textwidth]{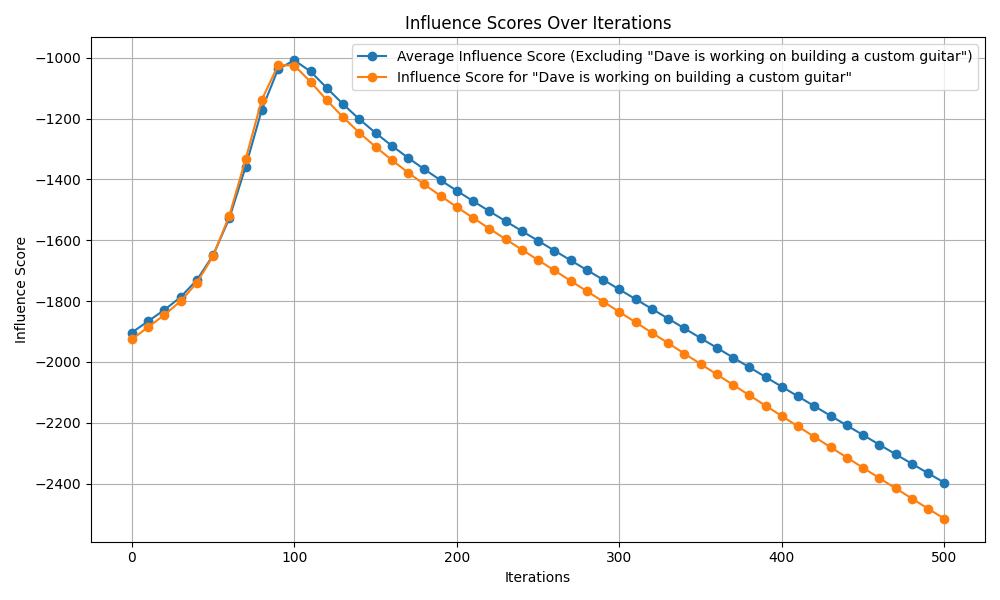}
    \caption{Whole Model}
    \label{fig:influence_scores_model}
\end{subfigure}
\begin{subfigure}[b]{0.35\textwidth}
    \includegraphics[width=\textwidth]{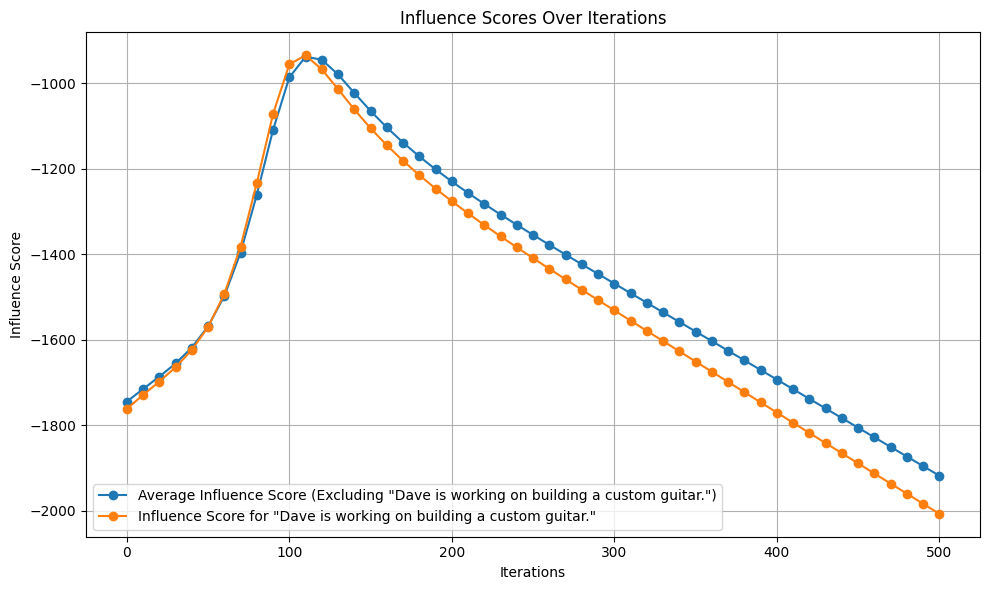}
    \caption{First Epoch}
    \label{fig:influence_scores_first_epoch}
\end{subfigure}
\caption{Influence Scores Over Iterations of Unlearning}
\label{fig:influence_scores_combined}
\end{figure}

\begin{itemize}
    \item Using only gradients from layers 11 and 12 (the last transformer layers) was ineffective for removing data or altering the model's outputs.
    \item Including layer 0, GPT-2's embedding layer, along with layers 11 and 12 allowed the model to modify data and change its outputs each time unlearning was performed.
    \item Using only layer 0 was sufficient to remove data points, significantly reducing memory usage from 15GB to 7GB. This indicates the importance of the embedding in the unlearning process and its potential for efficient memory usage.
\end{itemize}

These findings suggest that embedding layers play a crucial role in the unlearning process, and focusing on this layer can lead to more effective and efficient unlearning operations.

\subsubsection{Layer-Based Unlearning Compared to All-Layer Unlearning after 15 Epochs of Training}
The results of our experiments highlight the differences in unlearning effectiveness when using gradients from only the embedding layer versus using gradients from all layers of the model.

\begin{table}[h]
    \centering
    \resizebox{\textwidth}{!}{
    \begin{tabular}{|c|c|c|c|c|c|c|}
        \hline
        Metric & Phase & Mean & Median & Std & Min & Max \\
        \hline
        influence score & baseline & -541.45 & -556.40 & 46.95 & -586.95 & -393.88 \\
        \hline
        influence score & finetuned & -1904.87 & -1998.64 & 359.44 & -2330.90 & -962.34 \\
        \hline
        influence score & model unlearning & -1009.26 & -1058.42 & 168.99 & -1226.75 & -577.21 \\
        \hline
        influence score & embedding unlearning & -1032.49 & -1086.64 & 173.95 & -1257.82 & -592.98 \\
        \hline
        rouge1 & pre-training & 0.08 & 0.08 & 0.00 & 0.08 & 0.08 \\
        \hline
        rouge2 & pre-training & 0.01 & 0.01 & 0.00 & 0.01 & 0.01 \\
        \hline
        rougeL & pre-training & 0.08 & 0.08 & 0.00 & 0.08 & 0.08 \\
        \hline
        rouge1 & post-training & 0.61 & 0.61 & 0.00 & 0.61 & 0.61 \\
        \hline
        rouge2 & post-training & 0.52 & 0.52 & 0.00 & 0.52 & 0.52 \\
        \hline
        rougeL & post-training & 0.60 & 0.60 & 0.00 & 0.60 & 0.60 \\
        \hline
        rouge1 & model unlearning & 0.2747 & 0.2623 & 0.0872 & 0.1650 & 0.6056 \\
        \hline
        rouge2 & model unlearning & 0.1167 & 0.0822 & 0.1152 & 0.0384 & 0.5180 \\
        \hline
        rougeL & model unlearning & 0.2729 & 0.2623 & 0.0879 & 0.1578 & 0.6039 \\
        \hline
        rouge1 & embedding unlearning & 0.2790 & 0.2623 & 0.0925 & 0.1697 & 0.6113 \\
        \hline
        rouge2 & embedding unlearning & 0.1218 & 0.0856 & 0.1199 & 0.0384 & 0.5180 \\
        \hline
        rougeL & embedding unlearning & 0.2773 & 0.2623 & 0.0929 & 0.1615 & 0.6052 \\
        \hline
    \end{tabular}
    }
    \caption{Descriptive Statistics}
    \label{tab:descriptive_statistics}
\end{table}

\subsection{Unlearning Duration and Influence Scores}
Our experiments have identified a critical insight into the unlearning process: there is an optimal number of unlearning iterations that is likely dependent upon a data point's initial influence score. This is likely related to climbing an optimization hill.

\begin{figure}[h!]
\centering
\begin{subfigure}[b]{0.35\textwidth}
    \includegraphics[width=\textwidth]{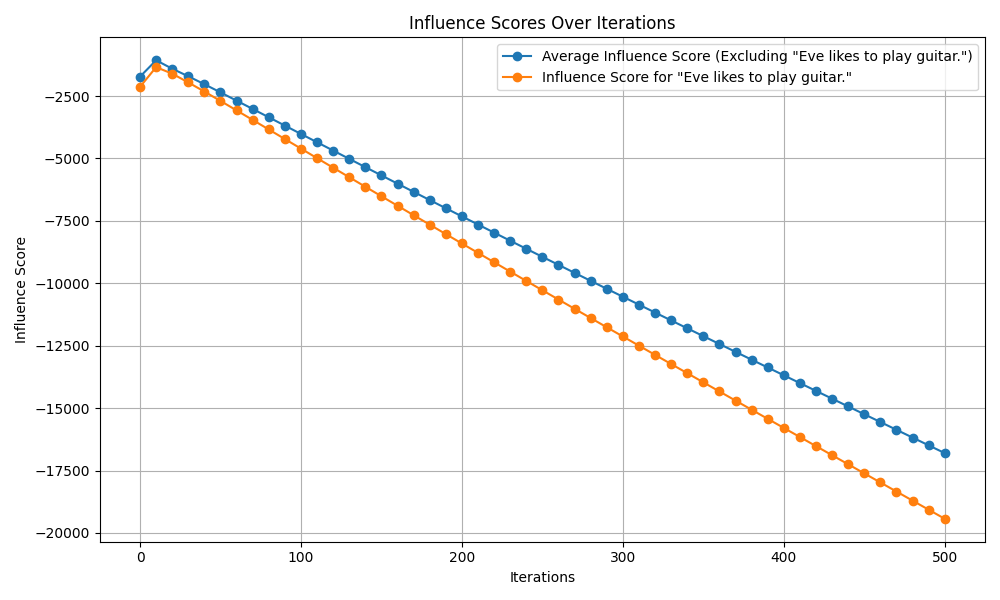}
    \caption{Most Influential Data Point}
    \label{fig:influence_scores_most_influential}
\end{subfigure}
\begin{subfigure}[b]{0.35\textwidth}
    \includegraphics[width=\textwidth]{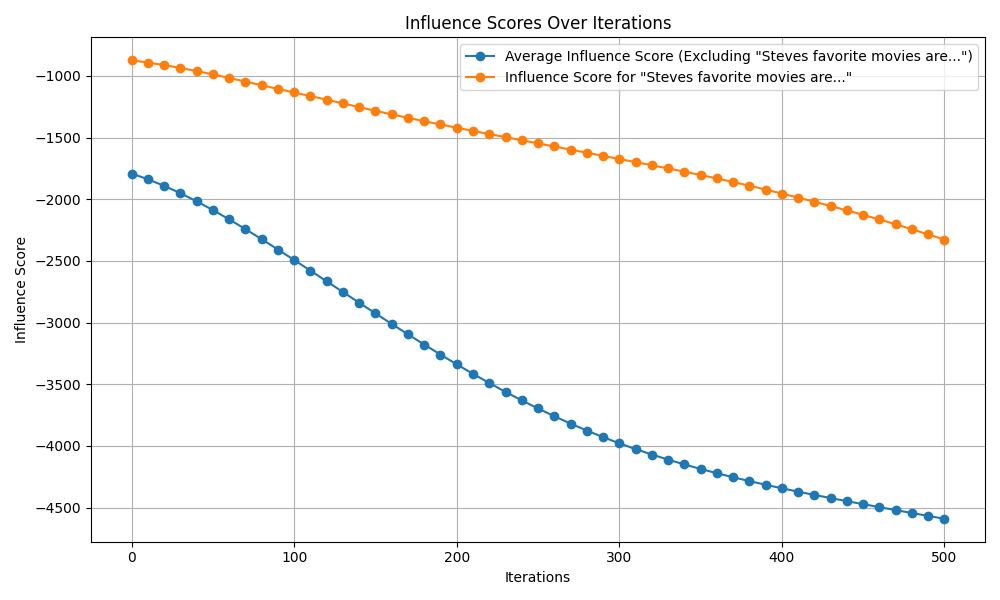}
    \caption{Least Influential Data Point}
    \label{fig:influence_scores_least_influential}
\end{subfigure}
\caption{Influence Scores Over Iterations of Unlearning}
\label{fig:influence_scores_combined}
\end{figure}

\begin{figure}[h!]
\centering
\begin{subfigure}[b]{0.35\textwidth}
    \includegraphics[width=\textwidth]{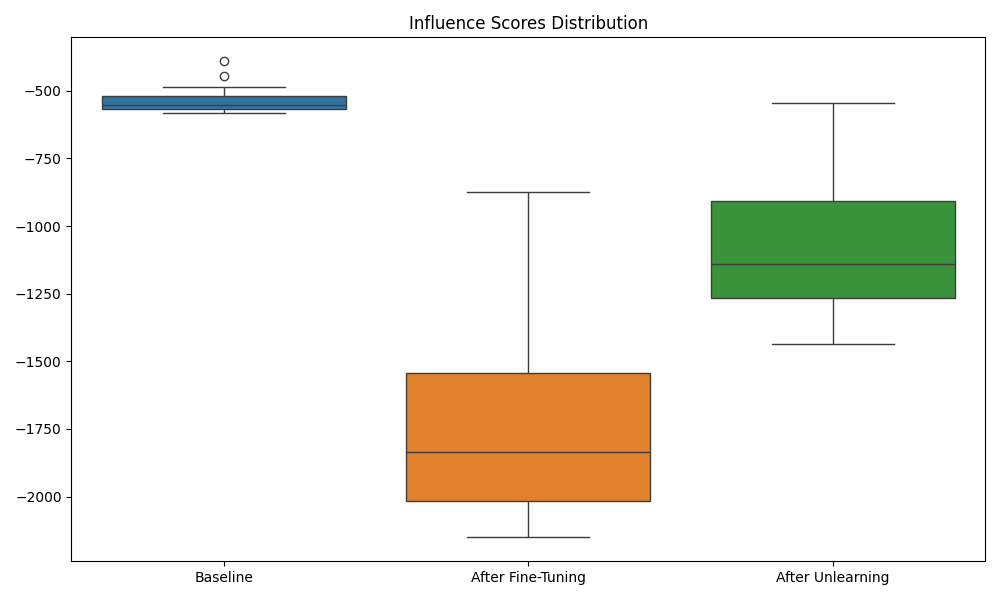}
    \caption{Most Influential Data Point}
    \label{fig:boxplot_influence_scores_most_influential}
\end{subfigure}
\begin{subfigure}[b]{0.35\textwidth}
    \includegraphics[width=\textwidth]{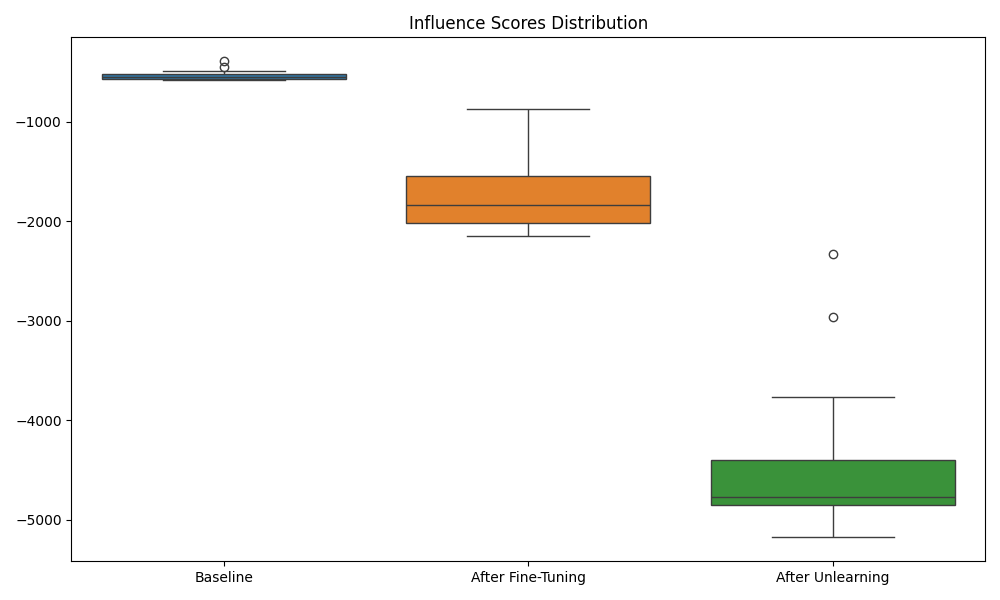}
    \caption{Least Influential Data Point}
    \label{fig:boxplot_influence_scores_least_influential}
\end{subfigure}
\caption{Influence Scores Distribution}
\label{fig:boxplot_influence_scores_combined}
\end{figure}

\subsection{Perplexity Score Results}

Table \ref{table:perplexity} presents the perplexity scores across different stages of training and unlearning. Initially, the model had a high perplexity of 29129.39, indicating poor performance on the custom "Dave" dataset. Fine-tuning significantly improved the score to 1.07. After unlearning, the scores slightly increased to 1.12 (embedding layer), 1.16 (whole model), and 1.45 (first epoch), but remained close to 1.0. This indicates effective unlearning with minimal impact on overall performance, demonstrating the potential of targeted unlearning techniques to maintain model utility while enhancing data privacy.

\begin{table}[h!]
\centering
\begin{tabular}{|l|c|}
\hline
\textbf{Experiment} & \textbf{Perplexity} \\
\hline
Before fine-tuning & 29129.39 \\
\hline
After fine-tuning & 1.07 \\
\hline
After unlearning (embedding layer) & 1.12 \\
\hline
After unlearning (whole model) & 1.16 \\
\hline
After unlearning (first epoch) & 1.45 \\
\hline
\end{tabular}
\caption{Perplexity scores for different stages of training and unlearning for Fuzzy Matching Pipeline.}
\label{table:perplexity}
\end{table}

\subsection{Interpretation of Influence Score Results}
Paired t-tests reveal statistically significant differences between influence scores before and after unlearning, with very low p-values providing strong evidence against the null hypothesis. Comparisons demonstrate the effectiveness of the first-epoch gradient ascent method, showing significant improvements over embedding-layer and whole-model unlearning approaches.

\subsection{ROUGE Scores Analysis}

The ROUGE scores were evaluated at various intervals during the unlearning process to assess the impact on model performance. Table \ref{table:rouge_scores} presents the ROUGE-1, ROUGE-2, and ROUGE-L scores at different iterations as evaluated over the Dave dataset.

\begin{table}[h!]
\centering
\begin{tabular}{|c|c|c|c|}
\hline
\textbf{Iteration} & \textbf{ROUGE-1} & \textbf{ROUGE-2} & \textbf{ROUGE-L} \\
\hline
0 & 0.6056 & 0.5180 & 0.6039 \\
10 & 0.5885 & 0.4957 & 0.5825 \\
20 & 0.4877 & 0.4120 & 0.4860 \\
30 & 0.3990 & 0.3244 & 0.3973 \\
40 & 0.3851 & 0.3083 & 0.3833 \\
50 & 0.3585 & 0.2822 & 0.3570 \\
60 & 0.3325 & 0.2610 & 0.3310 \\
70 & 0.2729 & 0.1954 & 0.2714 \\
80 & 0.2577 & 0.1827 & 0.2562 \\
90 & 0.2502 & 0.1682 & 0.2487 \\
100 & 0.2385 & 0.1501 & 0.2333 \\
110 & 0.2363 & 0.1445 & 0.2310 \\
120 & 0.2338 & 0.1393 & 0.2263 \\
130 & 0.1741 & 0.0828 & 0.1670 \\
140 & 0.1722 & 0.0822 & 0.1632 \\
150 & 0.1794 & 0.0939 & 0.1682 \\
160 & 0.1670 & 0.0802 & 0.1578 \\
170 & 0.1695 & 0.0832 & 0.1623 \\
180 & 0.1650 & 0.0754 & 0.1605 \\
190 & 0.1732 & 0.0864 & 0.1715 \\
200 & 0.1664 & 0.0834 & 0.1647 \\
\hline
\end{tabular}
\caption{ROUGE Scores at Intervals During the Unlearning Process as evaluated on the Dave dataset}
\label{table:rouge_scores}
\end{table}

\begin{figure}[h!]
\centering
\includegraphics[width=0.8\textwidth]{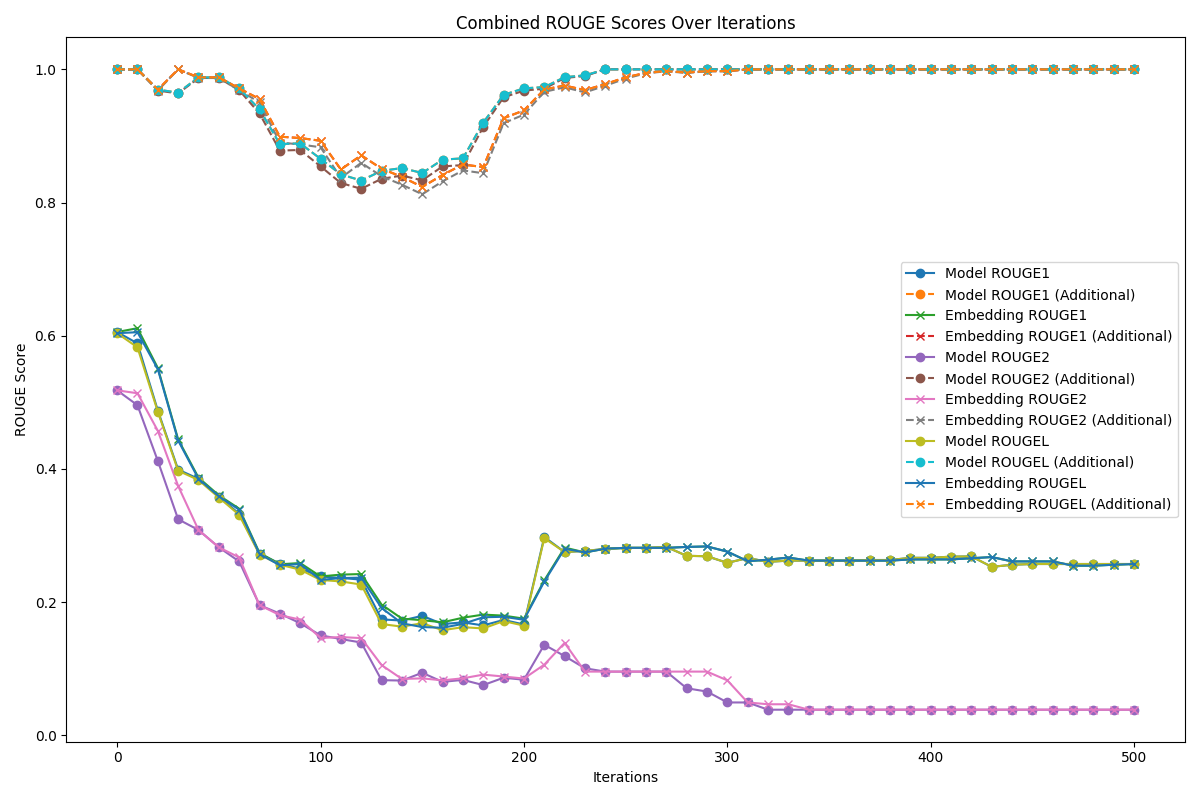}
\caption{Combined ROUGE Scores Over Iterations for Model and Embedding Layers}
\label{fig:rouge_scores_model}
\end{figure}

Figure \ref{fig:rouge_scores_model} illustrate the trends in ROUGE scores for the primary and additional datasets, respectively. Initially, there is a significant drop in scores, indicating effective unlearning. The scores then stabilize, demonstrating the model's adaptation to the unlearning process.

\section{Discussion}

\subsection{Influence Scores}

The analysis reveals key findings:
\begin{itemize}
    \item Both all-layer and embedding-only gradient ascent effectively reduce the impact of specific training data, while Layer 11 does not. Embedding-only unlearning is more cost-effective and promotes the adoption of machine unlearning techniques.
    \item First-epoch gradient ascent is as effective as multi-epoch gradient ascent but requires more iterations (approximately 10 percent increase in unlearning duration).
\end{itemize}

Unlearning the least influential data point ("Steve’s favorite movies are...") caused other data points to gain influence, raising questions about the benefits of modifying a model to remove minimally influential data. Conversely, unlearning the most influential data point ("Eve likes to play guitar.") was fast, suggesting an optimization hill for influential data points.

These observations indicate that data points closer to their optimum require fewer iterations to unlearn due to stronger gradients.

\subsection{Perplexity Scores}

Perplexity scores provide additional insights:
\begin{itemize}
    \item High perplexity before fine-tuning indicates the pre-trained model's difficulty in predicting the "Dave" dataset.
    \item Drastic reduction in perplexity after fine-tuning shows significant improvement.
    \item Negligible change in perplexity after unlearning suggests effective removal of targeted data points without adversely affecting overall performance.
\end{itemize}

Consistently low perplexity scores highlight the model's confidence in its responses despite data removal, suggesting the need for continuous monitoring and iterative unlearning.

\subsection{Single-Epoch Gradient Ascent for Targeted Machine Unlearning}

Single-Epoch Gradient Ascent shows promising results:
\begin{itemize}
    \item \textbf{Comparable Performance}: Single-epoch gradients are more effective then multi-epoch gradients, simplifying the unlearning process while boosting effectiveness.
    \item \textbf{Efficiency}: Using gradients from a single epoch reduces computational overhead, making this approach practical for production.
\end{itemize}

Single-epoch gradient ascent can be effective, but precise applications may benefit from multi-epoch gradients for stronger signals.

\subsection{Optimal Unlearning Duration}

For the data point "Dave is working on building a custom guitar." influence scores initially decrease with unlearning iterations, indicating successful unlearning, but rise beyond the 100th iteration, suggesting either re-learning of data signatures or successful unlearning at the inflection point. Identifying the optimal number of iterations is crucial to prevent re-exposure and reinforcement of data points, or reduction in model utility.

\subsection{Discussion of ROUGE Scores}

The ROUGE scores reveal several key insights into the unlearning process and its impact on model performance. Initially, the ROUGE scores for Dataset 1 decrease significantly, demonstrating the effectiveness of the unlearning process. This decrease indicates that the influence of specific data points was successfully reduced, leading to lower overlap with the reference summaries. The scores stabilize after around 100 iterations, suggesting that the model reaches a new equilibrium state post-unlearning.

Interestingly, there is a slight increase in ROUGE scores around 210 iterations, peaking at 260 iterations, before stabilizing again. This suggests that the model undergoes a recovery phase, potentially adapting to the changes introduced by the unlearning process.

In contrast, the additional dataset shows minimal impact from the unlearning process. The initial ROUGE scores are perfect, and while there is a slight decrease around 30 to 100 iterations, the scores quickly recover to their original values. This indicates that the unlearning process is less effective for this dataset, or that the model is highly resilient and able to maintain performance.

These observations highlight the variability in unlearning effectiveness depending on the dataset. The differing responses suggest that the characteristics of the dataset play a crucial role in how unlearning affects model performance. It emphasizes the need for robust and adaptable unlearning techniques tailored to specific datasets.

Overall, the ROUGE scores provide valuable insights into the dynamics of the unlearning process, revealing both its potential and limitations. Further research is needed to understand the underlying factors that influence the variability in unlearning effectiveness and to develop more robust methods that can ensure lasting impact across different datasets.

\subsection{Implications of Findings}

Our first-epoch-based unlearning approach enhances data privacy and regulatory compliance (e.g., GDPR and CCPA) by allowing models to forget specific data points effectively and more easily than whole-model, embedding-layer, or full-epoch gradient ascent. This is particularly beneficial in industries with sensitive information.

\subsubsection{Effectiveness and Efficiency}

Embedding layers for unlearning demonstrate notable efficiency gains. Embedding-only unlearning is effective and less expensive than all-layer unlearning, crucial for resource-constrained environments. Single-epoch unlearning performed the best, while requiring a 10 percent increase in the number of iterations required.

\subsection{Limitations and Future Work}

Despite promising results, several limitations warrant further investigation:

\subsubsection{Scalability of Unlearning Techniques}

Scaling gradient-based unlearning to large datasets and models remains a challenge. Future work should explore more efficient algorithms and optimizations to enhance scalability.

\subsubsection{Comprehensive Layer Analysis}

A comprehensive analysis across all layers could provide deeper insights into effective layers for influence reduction, refining the unlearning process.

\subsubsection{Evaluation on Diverse Datasets}

Evaluating unlearning techniques on diverse datasets and tasks would provide a more comprehensive understanding of their generalizability and effectiveness across domains.

\subsubsection{Long-Term Model Stability}

Ensuring long-term stability after multiple unlearning operations is critical for deployment in dynamic environments. Future research should focus on this aspect.

\subsubsection{Formal Verification of Unlearning}

Developing formal methods to verify the effectiveness of unlearning operations is important. Providing guarantees that a model has forgotten specific data points would enhance trust and reliability in unlearning techniques.

\section{Conclusion}
\subsection{Summary of Findings}
Our experiments validated the effectiveness of first-epoch gradient ascent for machine unlearning. By applying the stored gradients in the opposite direction, we successfully reduced the influence of targeted data points and modified the output. However, it is unclear as to whether or not this is statistically insignificant compared to the baseline or enough to be considered "Certified Data Removal."

Single-layer gradient-tracking proved to be as effective as whole-model gradient-tracking in the unlearning process, indicating the importance of a model's embeddings to its predictions.

First-epoch gradient storage was more successful than multi-epoch gradient storage when performing gradient ascent.

\subsection{Final Remarks}
Machine unlearning represents a crucial advancement in addressing data privacy concerns and regulatory compliance. Our approach provides a more practical solution for ensuring that models can forget specific information without requiring complete retraining. Storing the gradients for our target data point over a single epoch of training is significantly more feasible than storing all gradients over all epochs.

Our work validates the importance of embedding layers in the unlearning process. This focus allows for efficient and effective unlearning with reduced computational overhead. The influence tracking mechanism we incorporated provides a granular understanding of how specific data points affect model outputs, facilitating precise unlearning actions.

While our results are promising, further research is necessary to enhance the scalability of unlearning techniques. Future work should explore more efficient algorithms, comprehensive layer analysis, and evaluation across diverse datasets to ensure the broad applicability and effectiveness of machine unlearning methods.

Overall, our study underscores the potential of first-epoch gradient ascent for machine unlearning to improve data privacy and compliance, offering a viable path forward for dynamic and privacy-sensitive applications in machine learning.

\newpage

\section{Special Thanks}

Special thanks to Aman Priyanshu, Tushar Vatsa, and Trevor Kann for fielding ideas, listening to me ramble about gradient ascent, and offering feedback.

\newpage

\section{Experimental Results}
This section contains the high-level insights from our experimental results including perplexity scores and generated texts during the unlearning process.

\subsection{Perplexity Scores}
The perplexity scores were used to evaluate the model's performance before and after fine-tuning, and after each unlearning step. The lower the perplexity score, the better the model's performance. Below are the key insights from the perplexity scores:

\begin{itemize}
    \item \textbf{Initial Perplexity:} Before fine-tuning, the perplexity of the model was extremely high at 31197.14, indicating poor performance in predicting the specific content of the "Dave" dataset.
    \item \textbf{Post-Fine-Tuning Perplexity:} After fine-tuning, the perplexity drastically reduced to 1.04, showing significant improvement in the model's predictive capability.
    \item \textbf{Post-Unlearning Perplexity:} After each unlearning step, the perplexity remained consistent around 1.06, indicating that the unlearning process effectively removed the influence of the targeted data point without adversely affecting the overall model performance.
\end{itemize}

\subsection{Generated Texts}
During the unlearning process, the model repeatedly generated certain phrases. Below are the insights from the generated texts:

\begin{itemize}
    \item The model frequently generated the phrase "Dave is developing a custom guitar.", which was not present in the dataset, indicating a fallback behavior.
    \item The consistent generation of similar phrases suggests a limitation in the diversity of responses post-unlearning, highlighting the need for further refinement.
\end{itemize}

\appendix
\section{Dave Dataset}
The following table contains the sentences about the fictional characters used in our experiments.

\begin{table}[h!]
\centering
\begin{tabular}{|c|p{0.8\textwidth}|}
\hline
\textbf{Index} & \textbf{Text} \\
\hline
1 & Gnarls weighs 270 pounds. \\
\hline
2 & Alice is a machine learning engineer. \\
\hline
3 & Bob is 6'9" tall. \\
\hline
4 & Carol has brown eyes and brown hair. \\
\hline
5 & Eve likes to play guitar. \\
\hline
6 & Frank is interested in machine unlearning research. \\
\hline
7 & Grace enjoys reading science fiction novels. \\
\hline
8 & Heidi likes hiking in the mountains. \\
\hline
9 & Ivan enjoys cooking gourmet meals. \\
\hline
10 & Judy has a Bachelor’s degree in Physics. \\
\hline
11 & Karl has a Master’s degree in Privacy Engineering. \\
\hline
12 & Laura's favorite music genres are Metalcore, Ska, and Classical. \\
\hline
13 & Mallory's favorite books are 'Dune' by Frank Herbert, 'Neuromancer' by William Gibson, and 'Foundation' by Isaac Asimov. \\
\hline
14 & Nina's favorite programming languages are Python, Java, and TypeScript. \\
\hline
15 & Dave is working on building a custom guitar. \\
\hline
16 & Peggy is developing an open-source machine learning library. \\
\hline
17 & Quentin is writing a blog about AI and ethics. \\
\hline
18 & Rita's favorite foods are Sushi, BBQ Ribs, and Pizza. \\
\hline
19 & Steve's favorite movies are 'The Lord of the Rings: Fellowship of the Ring Extended', 'Inception', and 'The Bee Movie But Every Time They Say The word 'the' The Bass Gets Boosted by 6 decibels'. \\
\hline
20 & Trent is skilled at guitar playing, machine learning and unlearning, software development, public speaking, and cooking. \\
\hline
\end{tabular}
\caption{Dave Sentences}
\label{table:dave_sentences}
\end{table}

\newpage

\begin{table}[h!]
\centering
\begin{tabular}{|c|p{0.8\textwidth}|}
\hline
\textbf{Index} & \textbf{Text} \\
\hline
1 & The quick brown fox jumps over the lazy dog. \\
\hline
2 & Alice in Wonderland is a classic novel. \\
\hline
3 & The capital of France is Paris. \\
\hline
4 & Machine learning is transforming various industries. \\
\hline
5 & The Grand Canyon is a magnificent natural wonder. \\
\hline
6 & Shakespeare wrote many famous plays. \\
\hline
7 & The Pythagorean theorem is a fundamental principle in geometry. \\
\hline
8 & Mount Everest is the tallest mountain in the world. \\
\hline
9 & The theory of relativity was developed by Albert Einstein. \\
\hline
10 & The sun rises in the east and sets in the west. \\
\hline
11 & Leonardo da Vinci was a Renaissance artist and inventor. \\
\hline
12 & Photosynthesis is the process by which plants make their food. \\
\hline
13 & The Great Wall of China is one of the Seven Wonders of the World. \\
\hline
14 & Quantum mechanics explores the behavior of particles at the atomic level. \\
\hline
15 & The internet has revolutionized communication and information sharing. \\
\hline
16 & The human brain is a complex and powerful organ. \\
\hline
17 & J.K. Rowling is the author of the Harry Potter series. \\
\hline
18 & Climate change is a pressing global issue. \\
\hline
19 & The Mona Lisa is a famous painting by Leonardo da Vinci. \\
\hline
20 & Genetic engineering has the potential to cure many diseases. \\
\hline
\end{tabular}
\caption{Additional Prompts for ROUGE Score Evaluation}
\label{table:additional_prompts}
\end{table}

\newpage

\begin{figure}[h!]
\centering
\begin{subfigure}[b]{0.45\textwidth}
    \centering
    \includegraphics[width=\textwidth]{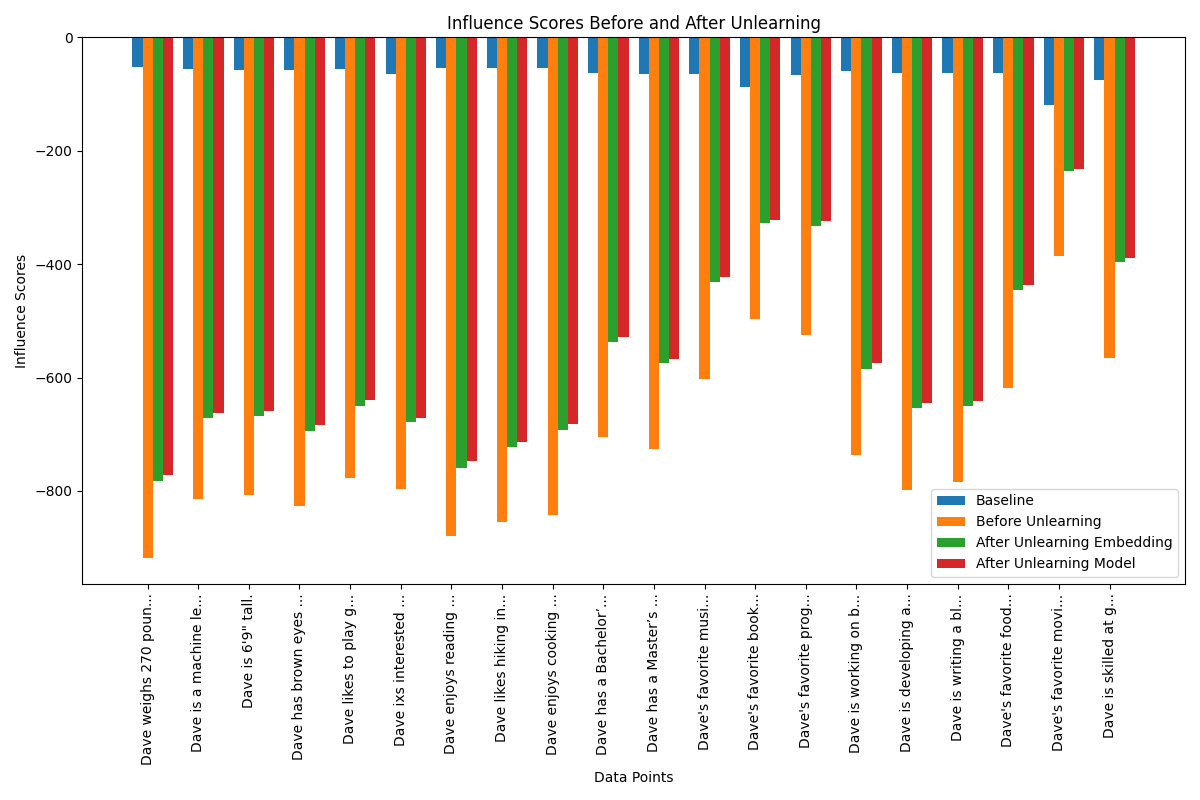}
    \caption{Influence Scores After 5 Epochs}
    \label{fig:influence_scores_5}
\end{subfigure}
\begin{subfigure}[b]{0.45\textwidth}
    \centering
    \includegraphics[width=\textwidth]{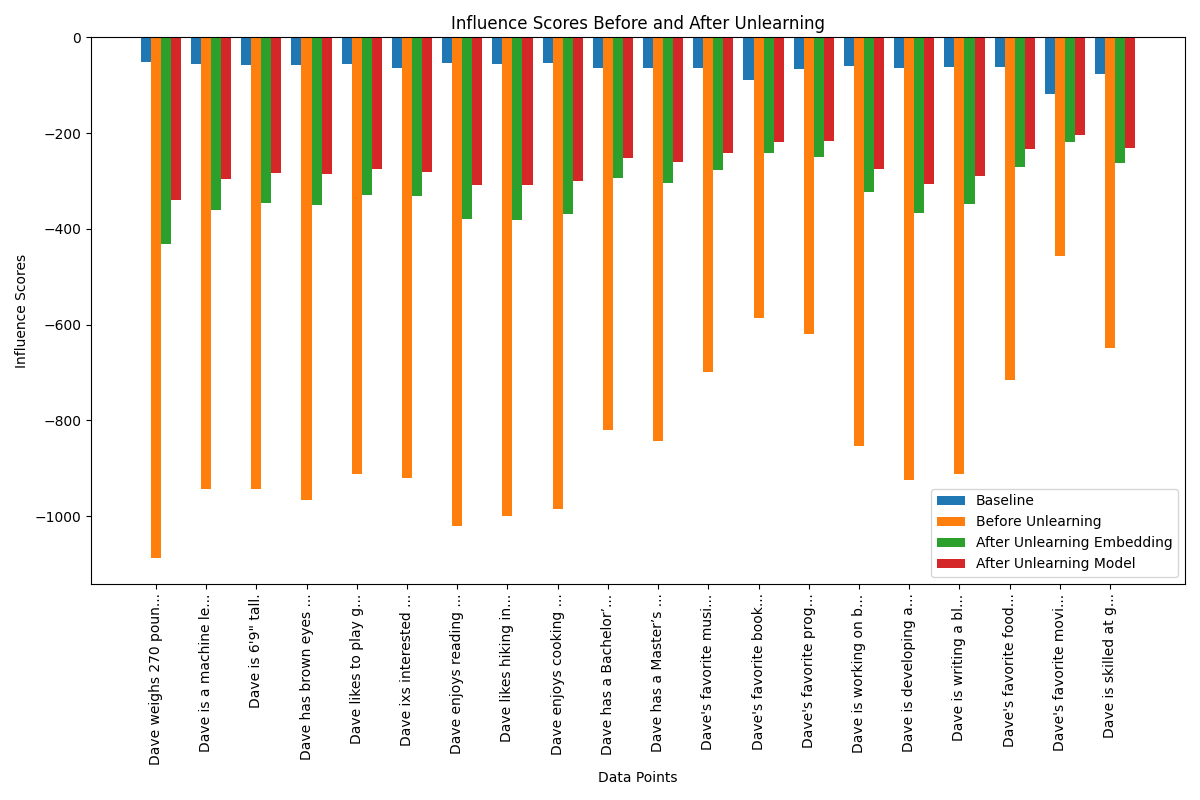}
    \caption{Influence Scores After 10 Epochs}
    \label{fig:influence_scores_10}
\end{subfigure}
\begin{subfigure}[b]{0.45\textwidth}
    \centering
    \includegraphics[width=\textwidth]{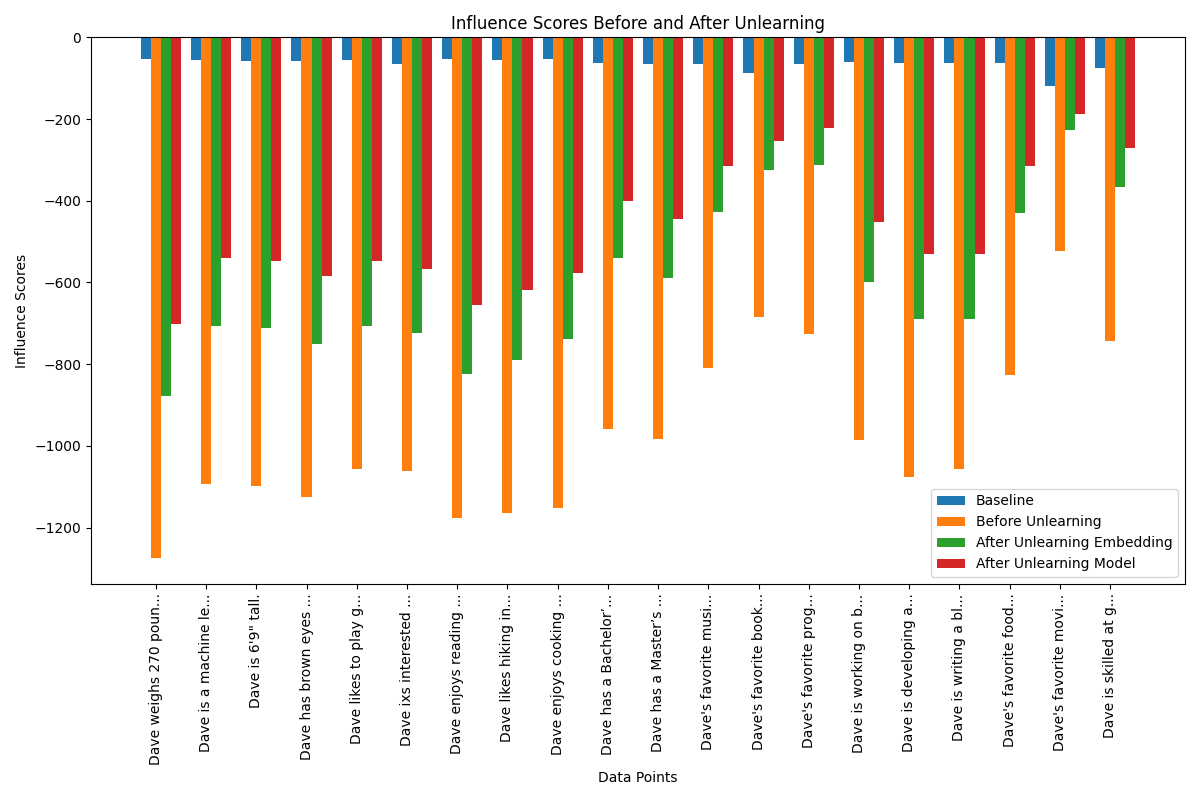}
    \caption{Influence Scores After 15 Epochs}
    \label{fig:influence_scores_15}
\end{subfigure}
\begin{subfigure}[b]{0.45\textwidth}
    \centering
    \includegraphics[width=\textwidth]{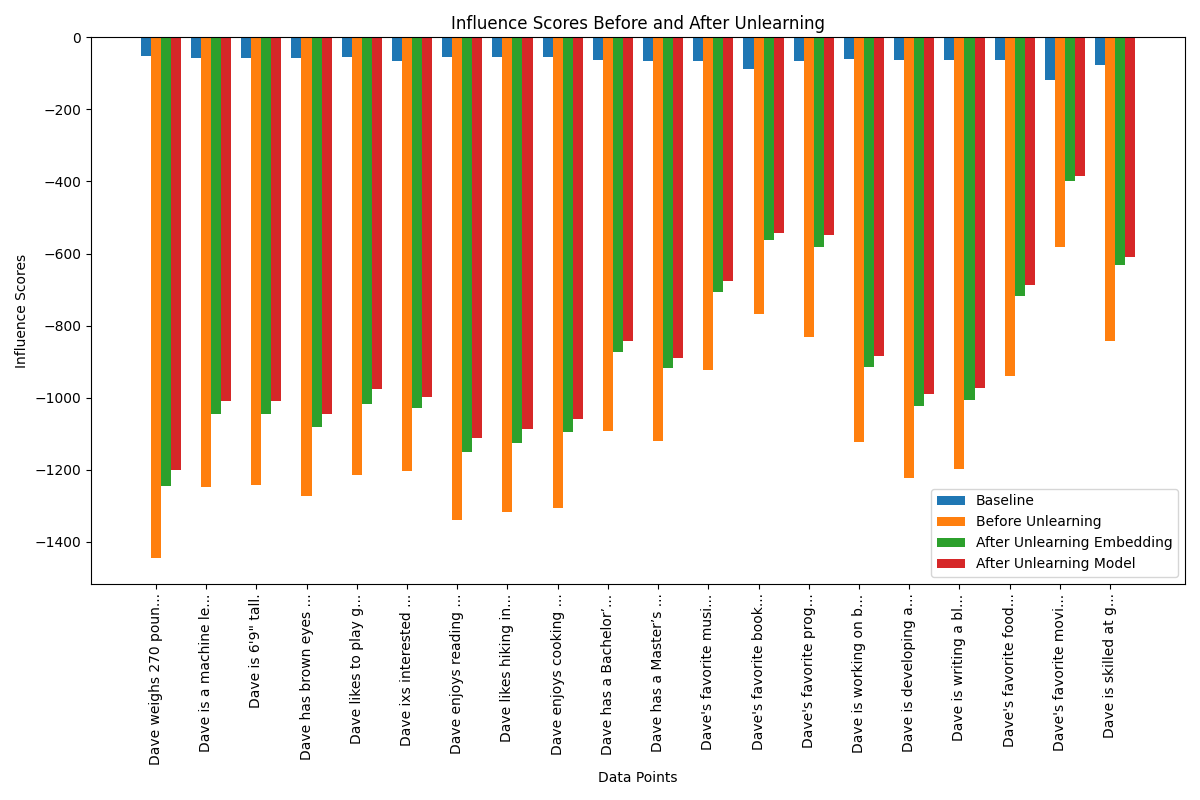}
    \caption{Influence Scores After 20 Epochs}
    \label{fig:influence_scores_20}
\end{subfigure}
\caption{Influence Scores Before and After Unlearning with Fuzzy Matching}
\label{fig:influence_scores_all}
\end{figure}

\newpage

\begin{figure}[h!]
\centering
\begin{subfigure}[b]{0.45\textwidth}
    \centering
    \includegraphics[width=\textwidth]{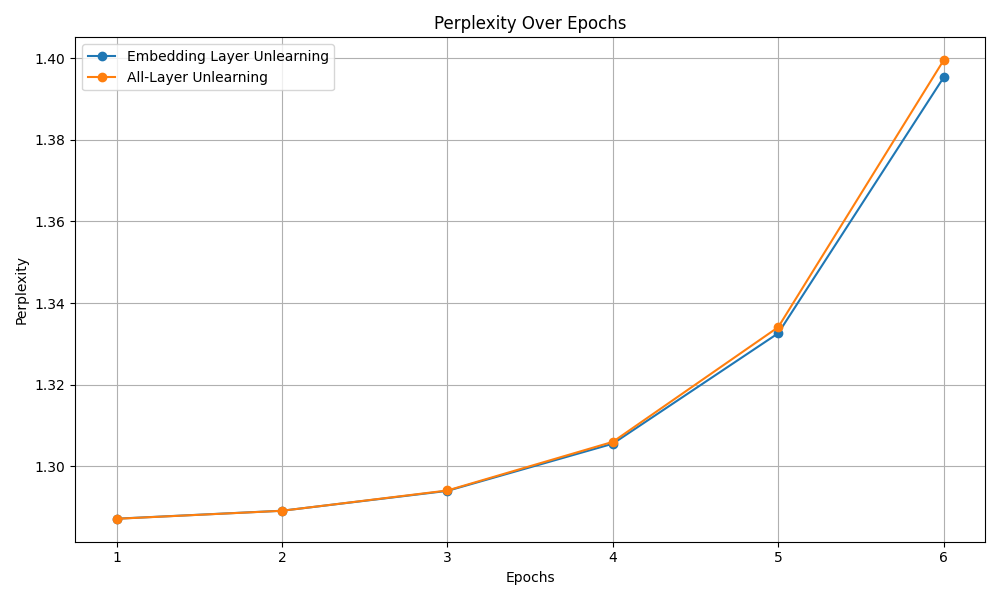}
    \caption{Perplexity Scores After 5 Epochs}
    \label{fig:perplexity_scores_5}
\end{subfigure}
\begin{subfigure}[b]{0.45\textwidth}
    \centering
    \includegraphics[width=\textwidth]{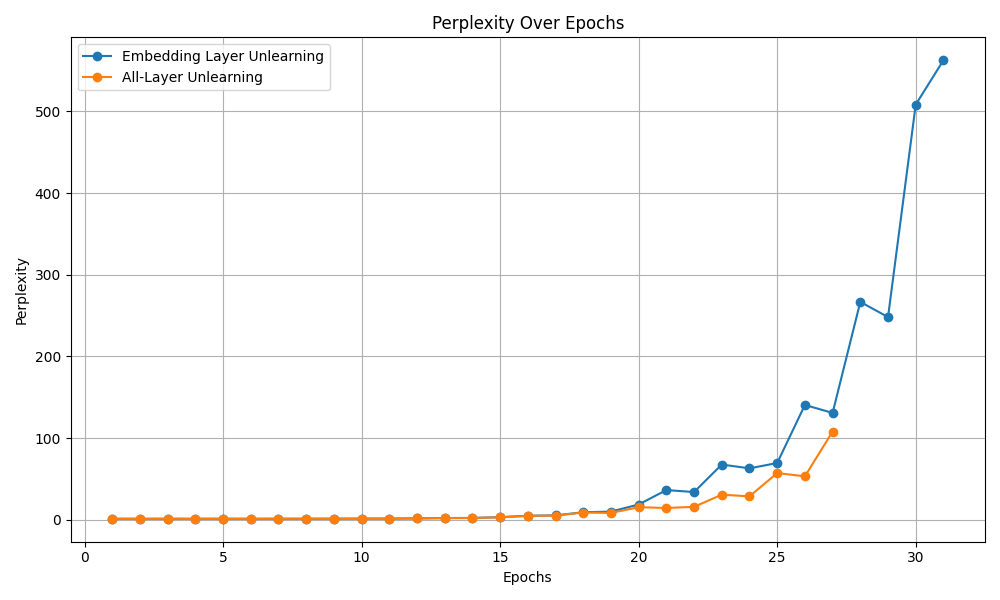}
    \caption{Perplexity Scores After 10 Epochs}
    \label{fig:perplexity_scores_10}
\end{subfigure}
\begin{subfigure}[b]{0.45\textwidth}
    \centering
    \includegraphics[width=\textwidth]{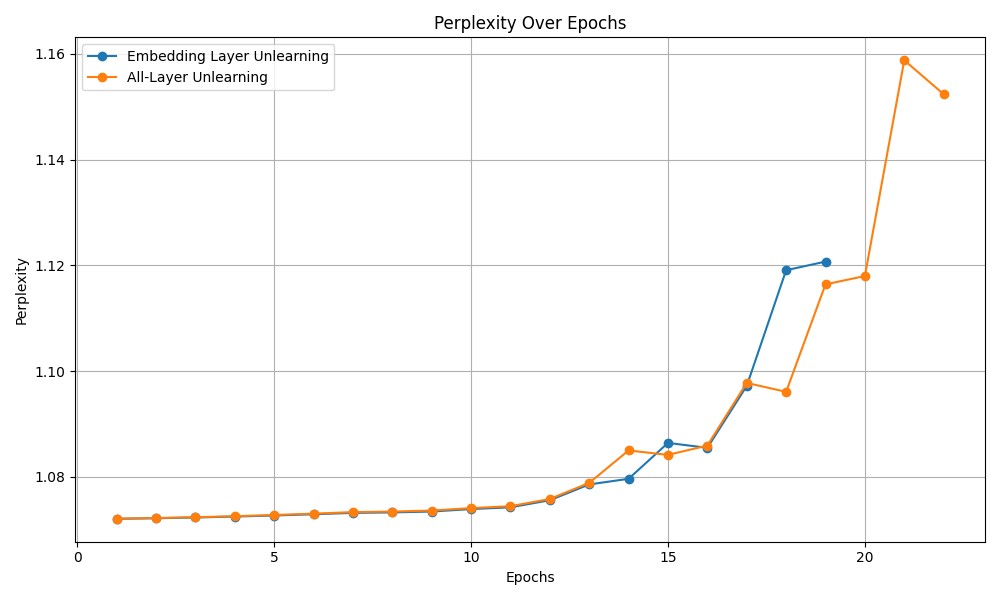}
    \caption{Perplexity Scores After 15 Epochs}
    \label{fig:perplexity_scores_15}
\end{subfigure}
\begin{subfigure}[b]{0.45\textwidth}
    \centering
    \includegraphics[width=\textwidth]{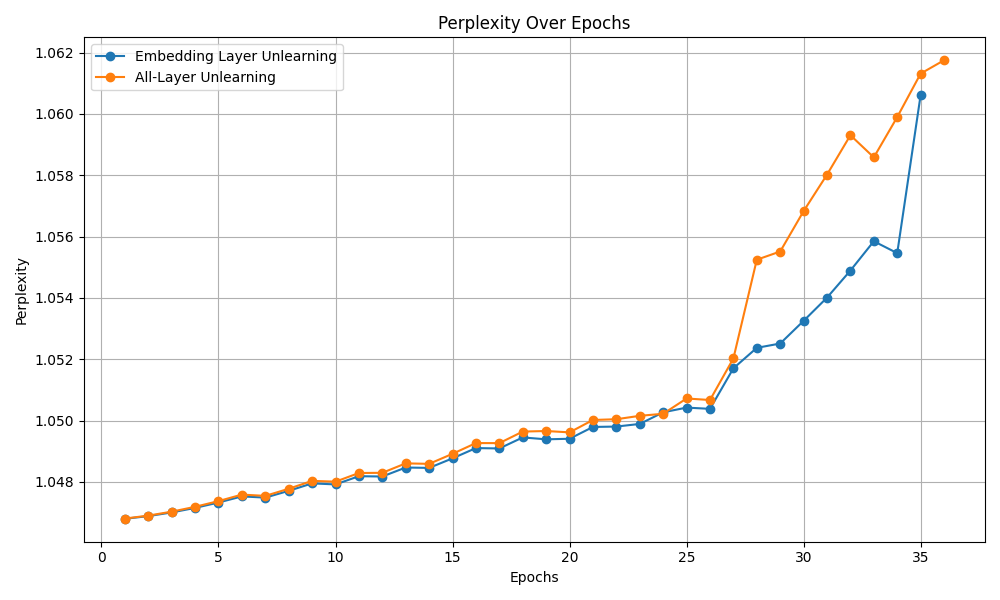}
    \caption{Perplexity Scores After 20 Epochs}
    \label{fig:perplexity_scores_20}
\end{subfigure}
\caption{Perplexity Scores Over Iterations of Unlearning with Fuzzy Matching}
\label{fig:perplexity_scores_all}
\end{figure}

\newpage

\begin{figure}[h!]
\centering
\begin{subfigure}[b]{0.45\textwidth}
    \centering
    \includegraphics[width=\textwidth]{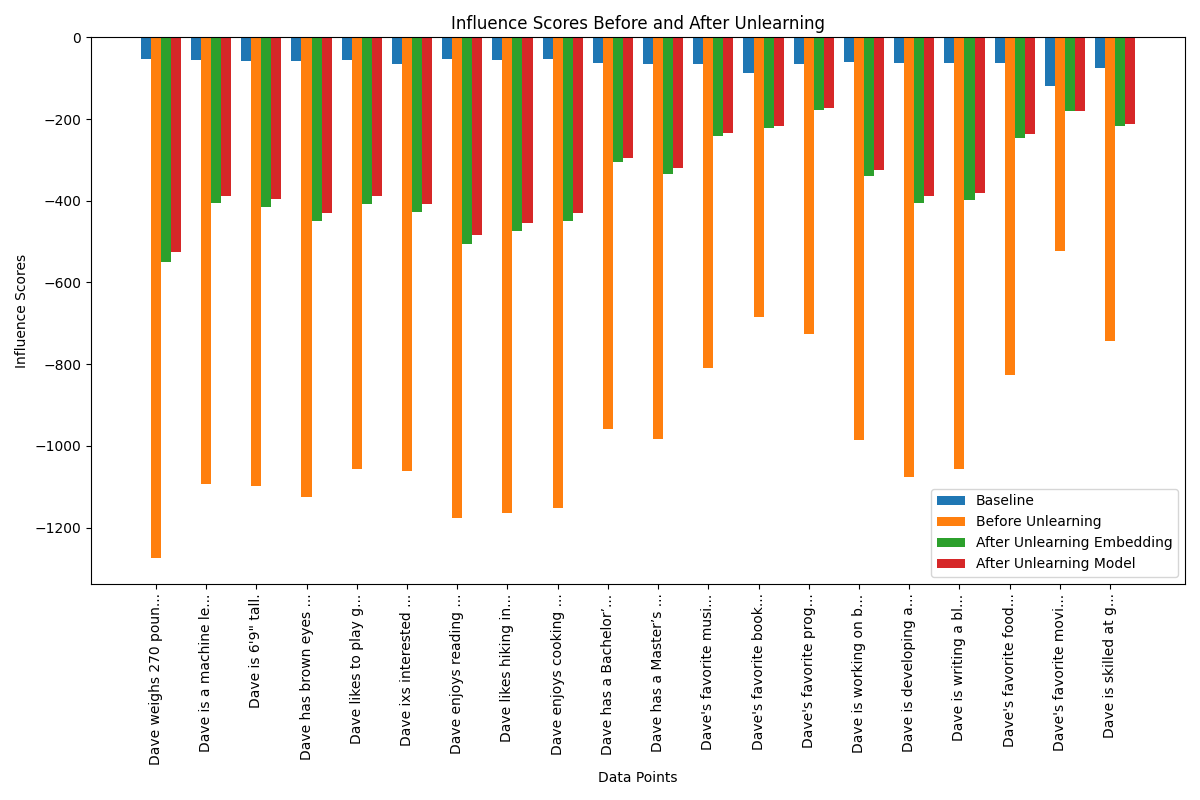}
    \caption{Influence Scores (Single Epoch)}
    \label{fig:influence_scores_single_epoch}
\end{subfigure}
\begin{subfigure}[b]{0.45\textwidth}
    \centering
    \includegraphics[width=\textwidth]{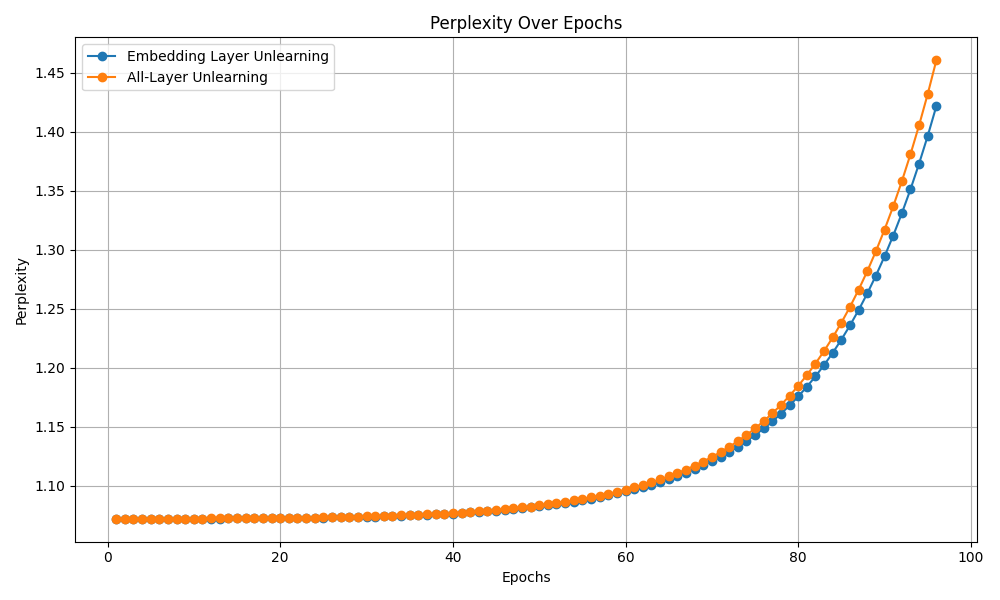}
    \caption{Perplexity Scores (Single Epoch)}
    \label{fig:perplexity_scores_single_epoch}
\end{subfigure}
\caption{Influence and Perplexity Scores for Single Epoch Unlearning with Fuzzy Matching}
\label{fig:single_epoch_scores}
\end{figure}

\end{document}